\documentclass[10pt, a4paper]{article}
\usepackage{lrec2014}
\usepackage[pdftex]{graphicx}
\usepackage{url}
\usepackage[utf8x]{inputenc}
\usepackage{tabularx}
\usepackage{booktabs}

\title{Getting Reliable Annotations for Sarcasm in Online Dialogues}

\name{Reid Swanson, Stephanie Lukin, Luke Eisenberg, Thomas Chase Corcoran and Marilyn A. Walker} 

\address{ University of California Santa Cruz \\
               Natural Language and Dialogue Systems Lab \\
               Computer Science Department \\ 
               reid@soe.ucsc.edu, maw@soe.ucsc.edu\\}

\abstract{
The language used in online forums differs in many ways from that of
traditional language resources such as news. One difference is the use
and frequency of nonliteral, subjective dialogue acts such as
sarcasm. Whether the aim is to
develop a theory of sarcasm in dialogue, or engineer automatic
methods for reliably detecting sarcasm, a major challenge is simply
the difficulty of getting enough reliably labelled examples.
In this paper we describe our work on methods for achieving highly
reliable sarcasm annotations from untrained annotators on Mechanical
Turk. We explore the use of a number of common statistical reliability
measures, such as Kappa, Karger's, Majority Class, and EM.
We show that more sophisticated measures do
not appear to yield better results for our data than simple measures such as
assuming that the correct label is the one that a majority of Turkers apply.
\\ \newline }

\begin{document}

\maketitleabstract

\section{Introduction}

The language used in online forums differs in many ways from that of
traditional language resources such as news. One difference is the use
and frequency of nonliteral, subjective dialogue acts such as sarcasm
as illustrated in Fig.~\ref{sarcasm-examples}.  Whether the aim is to
engineer automatic methods for reliably detecting sarcasm or to
further develop or test theories of sarcasm in dialogue
\cite{bryantfoxtree02,gibbs00}, a major challenge is simply the
difficulty of getting enough reliably labelled examples.

\begin{figure}[!htb]
\begin{center}
%\begin{scriptsize}
\begin{small}
\begin{tabular}{|p{2.35in}|c|}
\hline  
\bf Post Pair & \bf Category  \\ \hline \hline  
{\bf Q1:}  Not only have I undercut your snails. & \\
{\bf R1:} Oh No, everyone! Our Mollusca has been undercut! Whatever shall we do? :xbanghead emoticon-rolleyes &  SARC  \\ \hline  \hline
{\bf Q2:}  It is not possible for any animal to lay an egg that is "somewhat different but somewhat the same". DNA just doesn't work that way. &\\
{\bf R2:}  So you're a perfect clone of one your parents with zero copying errors? Amazing. &  SARC \\ \hline  \hline
{\bf Q3:} 
Did you know that in Siberia they are HAPPY that it's getting warmer up there? & \\
{\bf R3:}  Acutally global warming causes it to be hotter in the summer and colder in the winter, which is bad for any living thing. & NOT-SARC \\ \hline  \hline
{\bf Q4:} I'd love to know the exact questions asked.:) & \\
 
{\bf R4:}  Agreed.... there is nothing over at MORI about it. Perhaps Horizon will elaborate tonight. & NOT-SARC \\ \hline  
\end{tabular}
%\end{scriptsize}
\end{small}
\end{center}
\caption{\label{sarcasm-examples} Sarcasm Responses from Gold Label Set,
where Before category = After Category. Kappa, Kargers and Majority
all labelled the SARC examples with 1, and the NONSARC with 0. }
\end{figure}

In previous work, we released the Internet Argument
Corpus ({\bf IAC}), a large corpus of online social and political
dialogues \cite{Walkeretal12c}.  The topics in IAC cover a broad range of issues including
{\it evolution, gun control, abortion, gay marriage, existence of God,
  healthcare, communism vs. capitalism, death penalty, climate change}
and {\it marijuana legalization}.  The {\bf IAC} release includes Mechanical
Turk annotations from 5 to 7 annotators for dialogic categories 
of potential research interest, such as sarcasm,
disagreement and insults. These annotations reflect
Krippendorf Alpha scores ranging from .23 for sarcasm to .68 for
disagreement. In our 
previous work on automatic sarcasm classification,
henceforth referred to
as {\bf L\&W} \cite{LukinWalker13}, 
we  used a threshhold defined as {\it two annotators said it was sarcastic}
to define the subset of the corpus labeled as sarcastic.
The remainder of corpus was counted as non-sarcastic.
This definition leaves several open questions: 
\begin{itemize}
\item {\bf O1:} Did our definition of sarcastic vs. not sarcastic in our previous
research (threshhold of 2 annotators) help our reported results or hurt them? 
The not sarcastic category included posts that were labelled as sarcastic by 
zero or one annotator.
\item {\bf O2:} What makes it difficult to achieve high levels of agreement
for sarcasm annotation? Is sarcasm in the eye of the beholder, perhaps requiring a certain verbal subtlety to recognize?  Are there utterances that are truly ambiguous as to whether or not they are sarcastic? Is it simply
that there are unreliable workers on Mechanical Turk?
\item {\bf O3:} Do annotations of subjectivity in dialogue require more
annotators to achieve reliability than the NLP tasks of previous work
\cite{Snowetal08,callison2009fast}?
\end{itemize}

\begin{figure*}[!t]
\begin{center}
\begin{scriptsize}
%\begin{small}
\begin{tabular}{|p{2.95in}|c|c|c||c|c|c|c|}
\hline  
 & \multicolumn{2}{c||}{\bf Initial}  & \multicolumn{4}{c|}{\bf Final} \\
\bf Post Pair & \bf Karger's & \bf L\&W13  & \bf Kappa & \bf Majority & \bf EM & \bf CLASS \\ \hline \hline  

{\bf Q5:} The energy within everything is it's consciousness. &&&&&& \\
{\bf R5:}  Oh, I get it... Flashlights and I-Pods have souls and consciousness.
& AMBIG (.06)& SARC (5/5) & 1.0 & 1.0 & 1.0 & SARC \\ \hline  \hline
{\bf Q6:} 
Cuts off man's hand, lacerates his upper body, police say. ``
BALTIMORE - A Johns Hopkins University student armed with a samurai
sword killed a suspected burglar in a garage behind his off-campus
home early Tuesday, hours after someone broke in and stole
electronics. Some shocked neighbors said they heard bloodcurdling screams in an
area just blocks from the university. Police held the student, a
junior chemistry major who turns 21 on Sunday, for several hours, but
no charges were filed by early afternoon, said police spokesman Anthony Guglielmi. &&&&&& \\
{\bf R6:}  Oh, if we only had the requisite absolute civilian gun bans, sick things like this would never happen! Huh Brady? & AMBIG (-.003) & SARC (2/5) & 1.0 & 1.0 & 1.0 & SARC \\ \hline  \hline
{\bf Q7:}  You know, I think that you are rather an evil person, trying to create guilt where no guilt exists, where no guilt is necessary. Someone made a very important decision at a critical point in her life which made her life better and averted the consequences of a stupid mistake made by a youth. It was not an easy decision for her to make but she made it and it's well in the past, except for the people who want her to feel guilt and trauma about it.
&&&&& & \\
{\bf R7:}  Good lesson! Always take the easy way out to try and avoid consequences of actions. Instead of leaving a note on a car saying you accidently hit you...you should just casually drive away. Thanks Simone! & AMBIG (0.02)& SARC (4/7) & 1.0 & 1.0 & 1.0 & SARC \\ \hline  \hline
{\bf Q8:} There is nothing at all absurd about this take. The Gay agenda has been to normalize homosexuality and what better way to do it than to indoctrinate the young and desensitize them to the negatives of the lifestyle while they're still impressionable. That way they know that within 20 years they will have normalized this abnormal lifestyle.   &&&&& & \\
{\bf R8:} And this is a bad thing? & NON-SARC (-1.1) & NON-SARC (0/6) & .51 & .52 & .19 & AMBIG \\ \hline  \hline
{\bf Q9:} 
You are hot on the trail of the basis of our argument. I believe that there are three stage to the overall theory of evolution.  &&&&& & \\
{\bf R9:} I see. Then what you are claiming is that creationists are the ones who should define the vocabulary to be used by evolutionary biologists. Would it then be ok for atheists to define the vocabulary of Christianity? & AMBIG (0.04) & SARC (3/6) & .47 & .48 & .12 & AMBIG
\\ \hline  
\end{tabular}
\end{scriptsize}
%\end{small}
\end{center}
\caption{\label{hard-sarcasm-examples} 
Responses whose initial categorization varies by Karger's vs. {\bf L\&W} or
whose categorization differs after 25 annotations 
according to the reliability measures. Values for reliability measures are for the SARC category.Values for NON-SARC are the difference to 1.0. Only the Responses {\bf R} utterances
are labelled for sarcasm. }
\end{figure*}

Questions {\bf O2} and {\bf O3} raise
the issue that the nature of the linguistic phenomenon and its instantiation
in a particular genre should also be considered when considering reliability
measures.  Most annotation tasks in computational linguistics over
the last 20 years have focused on problems that have an objective
ground truth, such as part-of-speech and syntactic analysis.  However,
recently there has been a growing interest in other linguistic
phenomena that are more subjective in nature, such as
detecting sarcasm, opinions, or emotions in online discourse.  Not
only is it difficult to disambiguate the interpretation of the speaker
after the fact, but in conversation, it is also possible that the
speaker is being intentionally ambiguous.  For example, R8 in
Fig.~\ref{hard-sarcasm-examples} might have been deliberately
constructed by the speaker to be ambiguous, in the same way that
indirect speech acts may be constructed to be ambiguous
\cite{BrownLevinson87,Levinson81,Levinson85}. To our knowledge this
aspect of sarcasm has not been discussed by previous work on its
automatic recognition.

Here we aim to further explore the issues in achieving reliable
sarcasm annotations from untrained annotators on Mechanical Turk.  We
apply different reliability measures to the same data, including
majority class, Karger's, Kappa and EM
\cite{Dieugenioglass04,DawidSkene79,Kargeretal11}. In some cases these
measures make different predictions as exemplified by
the examples in Fig.~\ref{hard-sarcasm-examples} which are initially ambiguous
according to Karger's (Q5, Q6, Q7 and Q9), but which the L\&W
threshhold would classify as SARC (see column L\&W13).

Previous work on sarcasm in Twitter has mainly assumed that the
user-generated \#sarcasm tag reliably identifies sarcastic utterances
\cite{gonzalezetal11,davidovetal10}, although a recent study by
\cite{Riloffetal13} found that only 45\% of the utterances tagged as
\#sarcasm in a large corpus of Twitter utterances were judged by human
annotators to be sarcastic without any prior context.
\cite{Filatova12} reports a crowdsourcing study for identifying
sarcasm in product reviews on Amazon, and describes a procedure
for achieving a corpus with highly reliable labels, but does not
actually report reliability statistics. Moreover
because product reviews are much longer than either tweets or utterances
in online forums, it would be reasonable to assume that
sarcastic utterances in reviews are much less context dependent, and that
in general the length of the reviews would increase reliability for
sarcasm annotation.

Thus, we report the first study of the issues
involved with achieving high reliability labels for sarcasm in
online dialogue. 
Interestingly, our results suggest that for our data,
more sophisticated measures of Turker or annotator
reliability do not appear to yield clearly better results than
``naive'' measures, e.g. thresholding on number of annotators,
or assuming the correct label is the majority label.

\section{Sarcasm Corpus and Models of Reliability}
\label{reliability-meas-sec}

\subsection{Sarcasm Corpus}

The initial IAC annotation involved 10,003 
 Quote-Response ({\bf Q-R}) pairs where Mechanical
Turkers were shown seven Q-R pairs
and asked to judge whether the response was sarcastic or not. Example
{\bf Q-R} posts are in Fig.~\ref{sarcasm-examples} and
Fig.~\ref{hard-sarcasm-examples}.  Turkers were not given additional
definitions of the meaning of sarcasm, e.g. we let Turkers use their
native intuitions about what it means for a post to be sarcastic,
since previous work suggests that non-specialists tend to collapse all
forms of verbal irony under the term sarcastic
\cite{bryantfoxtree02,gibbs00}.

For each of these 10,003 Q-R pairs we collected annotations from
5 to 7 Mechanical Turkers. Previous work had tested the reliability
of Mechanical Turker annotations as compared to expertly trained
annotators for five different NLP tasks: affect
recognition, word similarity, recognizing
textual entailment, event temporal ordering,
and word sense disambiguation \cite{Snowetal08}.
In this work, the number of annotators required to match the expert
annotators ranged from 2 for labelling the anger affect,
to 10 for the textual entailment task. This was the basis
for collecting 7 annotations per item  for IAC.

The threshhold used by {\bf L\&W} on IAC to identify
the correct label ({\bf SARC} vs. {\bf NON-SARC}) was that {\it two
  annotators had to have annotated the post as sarcastic}. 
This threshhold was based on our observations of a
sample of the corpus.  Fig.~\ref{hard-sarcasm-examples} provides
examples of a number of posts along with the number of annotators in
the original IAC who marked the response (column
labelled {\bf L\&W}).  {\bf L\&W} labelled all of the remaining posts, including those marked by one annotator as
sarcastic, as non-sarcastic. Question {\bf O1} aims to identify
any issues with this definition.

\subsection{Reliability Measures}

There are at least three different types of reliability affecting
annotation studies and many factors that affect them. These are
reliability of the {\bf task}, {\bf annotators}, and {\bf item}.

The issue of the {\bf task} concerns whether the annotation task is well
defined and the process can be reliably reproduced.  This is usually
assessed using a measure of inter-rater reliability, such as Cohen's
$\kappa$ or Krippendorff's $\alpha$.  In our previous work on
collecting annotations for sarcastic responses we obtained an $\alpha$
of 0.23.  In this study, with a much larger pool of annotators and
some explicitly non-ambiguous cases we found a slightly higher
agreement ($\alpha$ = 0.387).  This is generally considered low for
objective annotation tasks, but it is unclear what this means for
subjective tasks such as sarcasm annotation. 

The issue of the {\bf annotators} concerns the quality of naive annotators
crowdsourced from Mechanical Turk. We would like to assess the
reliability of individual annotators, and several of the reliability
models discussed below are explicitly concerned with developing methods of weighting the reliability of annotators.  This is related to the general
problem of trust and reputation management systems.

The final issue concerns a measure of confidence or reliability
for the individual labeled {\bf items} in a gold standard corpus, or indeed
in the corpus as a whole.  Understanding which cases are difficult for
people to recognize supports a better analysis of automatic classification
errors, i.e. does it make worse
predictions on low confidence items \cite{LouisNenkova11}?

There are many different factors that can influence these facets of
reliability.  In any annotation effort the clarity of instructions and
annotation guidelines plays a large role in the consistency of the
results that are obtained.  The instructions should be clear and
thoroughly explain the requirements, however, they should also be
suitable for the target audience.  When using naive annotators
on Mechanical Turk, requesters generally strive to construct tasks
so that they are simple and straightforward, rather than 
providing detailed technical descriptions of the task.
We now describe different techniques for measuring reliability.

\noindent{\bf Simple Majority Voting.}  A common method for
determining which label to use as a gold standard is to use the one
selected by the majority of annotators.  Using an odd number of
annotators on a binary task will guarantee a single unique gold
standard label, otherwise selecting a random item from the shared
majority is a common way to break ties.  Majority voting is often used
because it is simple to implement and has been shown to be effective
on other tasks with a large number of annotators.  This approach can
also be effective when all of the annotators are well trained and
likely to be nearly equally reliable.

This method is problematic for naive annotators that are typically
used in crowdsourced annotation.  The skill-level, trust and
reliability of annotators on a crowdsourcing platform, such as AMT,
varies widely.  It can also be challenging to
broadly categorize annotators on these platforms, for example by using
their overall acceptance rate, because good performance on one task
does not always translate to high levels of reliability on a different task.  Moreover, simple majority voting does not take the
different levels of reliability for individual annotators into account
and can lead to low skill or malicious annotators  influencing the
final result too strongly.

Majority voting does not give a direct measure of reliability for an
individual annotator.  However, it is easy to construct one by
calculating the percentage of times an annotator agreed with the
majority.  By keeping the fraction of votes for each label it is also
possible to use this method to estimate the confidence in the
assignment of each label.
\cite{Kargeretal11} prove that
majority voting is sub-optimal and can be significantly improved upon.

% TODO is it worth running experiments using Majority Weighted Voting (Callison-Burch)
\noindent{\bf Kappa Weighted Voting.} An extension to simple majority
voting is to weight each vote by some estimate of the reliability of
the annotator who provided that label.  Our first approach weights
each annotator's vote using a standard inter-rater reliability
measure.  Annotators who have high agreement with other annotators
should be considered more reliable and given more weight than those
who do not.  Cohen's $\kappa$ is a standard measure of inter-rater
reliability that estimates the agreement of the annotated labels
adjusted for chance; this was used for the basis of weighting our votes
in this approach.

Cohen's $\kappa$ is usually computed between 2 annotators to produce a
measure of reliability of the data.  To produce a measure of
reliability of the annotator we calculated the average pairwise
$\kappa$ between that rater and all other raters for which there were
at least 10 annotations in common (1 HIT) over all the available data.
The final gold standard labels were produced by multiplying each
annotator's response ($\pm$ 1) by their $\kappa$ score and summing the
total.

This approach is conceptually similar to the non-expert weighted
voting method by Callison-Burch~\cite{callison2009fast}, which
weighted each vote by how often the annotator agreed with the majority
over the entire dataset.  Callison-Burch showed that this method
worked quite well for evaluating translation quality, reaching near
expert levels of performance with only 5 annotators.  The $\kappa$
value used as the annotator weight can also be used to assess the
reliability of individual annotators.

One drawback to this method is that tasks on Mechanical Turk 
tend to be structured such that most annotators only perform a
small fraction of the total number of annotations that are available.
This creates a very weakly connected network of annotators, which  may be
unfairly lowering the reliability of high quality annotators (or vice
versa).  For example, a highly reliable annotator might share
annotations in common with 5 other annotators.  If 3 of those
annotators happen to be very unreliable, then it will also appear that
the highly reliable annotator is  of dubious quality.

\noindent{\bf Karger's Algorithm.} Karger's
algorithm~\cite{Kargeretal11} is an iterative message passing
algorithm that attempts to address this issue.  This algorithm models
the problem as a weighted bipartite graph with nodes representing the
annotators and tasks as distinct sets.  An edge is created between an
annotator node and a task node if that annotator provided a label for
that item.  The weights of the nodes can be positive or negative and
can be thought of as the reliability of the corresponding annotator
and individual annotation task.  Positive values for annotator nodes
indicate a belief that their answers correlate with the true label,
negative values indicate their answers correlate with the opposite
label and values near zero indicate we are uncertain about their
contributions.  Similarly, positive values for task nodes indicate a
belief that the true label of the item is $+1$ (i.e., sarcastic),
while negative values suggest the true label is $-1$ (i.e., not
sarcastic).  The magnitude of the weights gives a relative measure of
how strong our belief is.

%% POSSIBLY FIGURE TO ILLUSTRATE

The weights of each annotator are initialized to 1.  The algorithm
begins by passing a message from each annotator node to all connected
task nodes.  The message is constructed by multiplying the annotator's
current reliability estimate (i.e., the node weight) by the label
($\pm$1) that the annotator provided for the receiving task node.  The
messages entering each task node are summed and become the weight of
that node.  On the second, step messages are passed back from the task
node to each annotator they are connected to.  The message is
constructed by taking the weight of the task node and subtracting out
the contribution of the receiving annotator, then multiplying the
adjusted weight by the annotator's original label.  This has the effect
of passing back a large positive value if many other reliable
annotators labeled the task with the same label and vice versa.  These
messages are summed and then normalized to become the new
weight/reliability for the annotator node.  The
process repeats until convergence or a fixed number of iterations
(e.g, 10).

This algorithm has the advantage that it can detect malicious raters who consistently provide incorrect labels and use this information to bolster the confidence of the actual true label.
The algorithm we used is only appropriate for binary labeled data, which is fine for our purposes, although a modification has been proposed to handle multi-label data.

However, Liu et al.~\cite{liu_variational_2012} have argued that Karger's method suffers several shortcomings that cause it to perform poorly on several real world NLP annotation tasks, such as the Recognizing Textual Entailment challenge~\cite{dagan_pascal_2006}.
Liu et al. demonstrate if the problem is framed as inference in a graphical model using a Beta prior then performance can be substantially improved for all the problems they investigated.
We also noticed in our work that Karger's method has problems dealing with annotation tasks when there is low overlap between annotators and a large variability in the number of items each annotator labels.
In these cases Karger's tends to conflate reliability with productivity.
In the future we will investigate the variational methods proposed by Liu et al.

% TODO say something about how this has problems when you have some annotators that do a lot of annotation and some that do not.

\noindent{\bf Dawid \& Skene's EM.}
An alternative probabilistic method was developed by Dawid and Skene~\cite{DawidSkene79}.
In this approach we estimate the probability that a task will be given an observed label $L_{o}$ by an annotator $A$, where
we have a true label $L_{t}$ given our data.
This can be written as:
\begin{equation}
 p(L_{o}|A,L_{t}) p(L_{t})
\end{equation}
where $p(L_{o}|A,L_{t})$ can be thought of as the annotator error rates and $p(L_{t})$ as the prior likelihood of the true labels.
Assuming the data is i.i.d then the probability of the true labels is given by
\begin{equation}
 \prod_{k=1}^{|A|}\prod_{l=1}^{|L|}p(l_{o}=l|a=k,l_{t}=j)p(l_{t}=j)
\end{equation}
The error rates and priors are then estimated using the Expectation-Maximization algorithm using the maximum likelihood estimation of the probabilities.
After convergence (or a fixed number of iterations) the estimated parameters are used to make a final determination about the true label of each task.
This algorithm has the advantage of providing probabilistic interpretations of the results and handles multi-category data without modification.

\section{Experimental Method}
\label{method-sec}

\begin{figure}[h!]
\begin{center}
\includegraphics[width=3.25in]{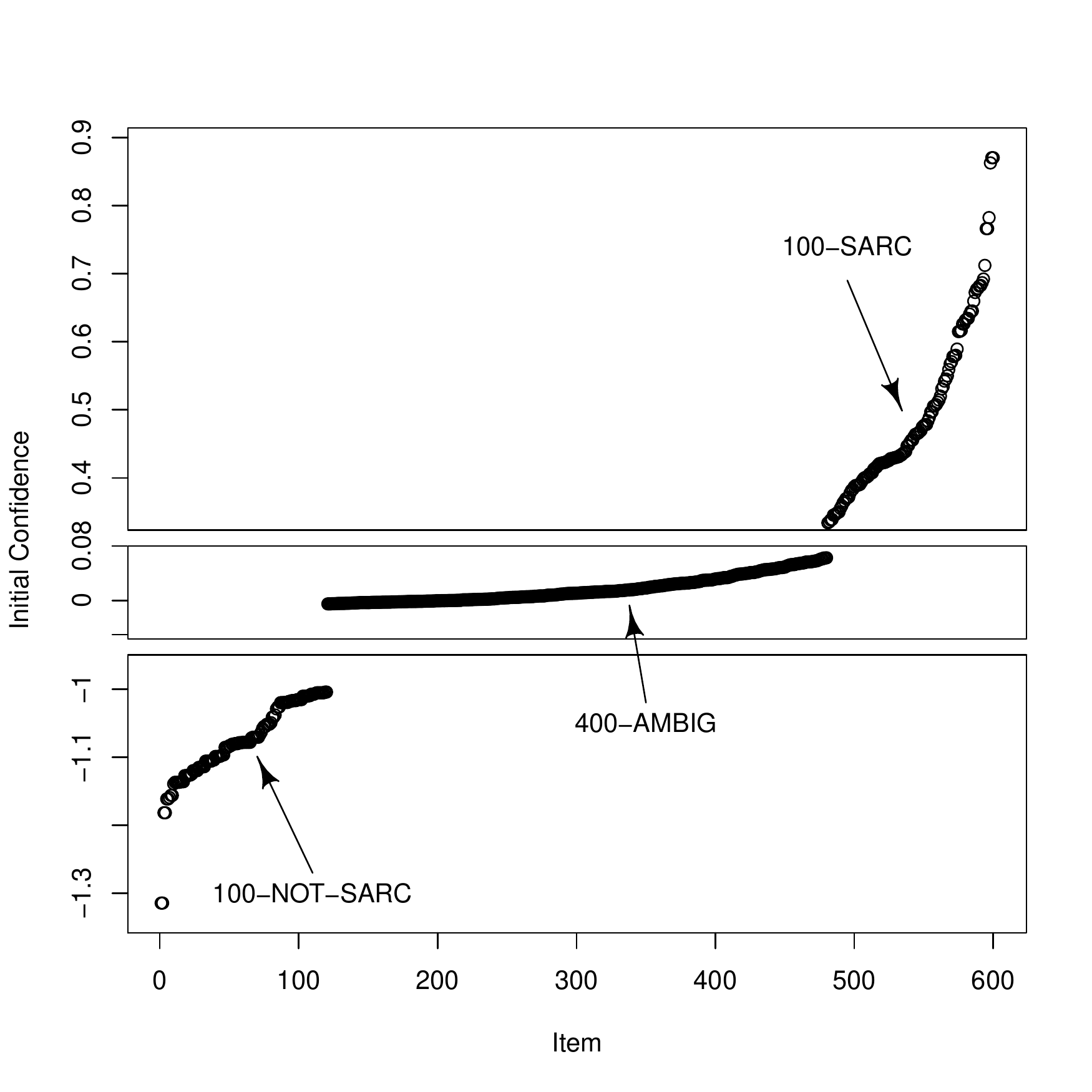}
\vspace{-0.2in}
\caption{\label{initial-distrib-fig} 600 utterances selected
from IAC according to Karger's confidence of labels for non-sarcastic (-1) and
sarcastic (1), from 5-7 annotations as distributed in IAC.}
\end{center}
\end{figure}

We begin by applying Karger's model to the 5-7 annotations released
with the {\bf IAC}, and calculate a confidence score for sarcastic
vs. not-sarcastic over the 10K utterances in the {\bf IAC}.  From
these 10K utterances, we then selected a set of 600 utterances for the
experiments reported here.  Fig.~\ref{initial-distrib-fig} shows the
Karger confidences for these 600 utterances.  We select 100 utterances
deemed to be reliably non-sarcastic ({\bf 100-NOT-SARC}), 100 judged
to be reliably sarcastic ({\bf 100-SARC}), and 400 utterances that
were not clearly either sarcastic or not sarcastic ({\bf
  400-AMBIG}). Of these 400, we select 300 that were very slightly
more likely to be sarcastic, and 100 that were very slightly more
likely to be not-sarcastic.
To select the ambiguous items we sorted the items by their Karger score and found the item with the score closest to zero.
We then selected the 300 items above this point (max score of 0.63) and 100 items below this point (min score of -0.005).

\begin{figure}[h!]
\begin{center}
\includegraphics[width=3.25in]{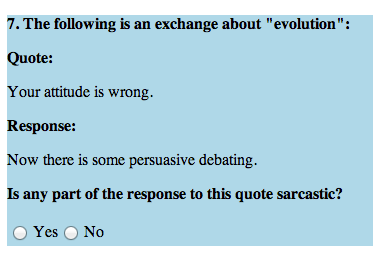}
\vspace{-0.2in}
\caption{\label{hit-fig} Sample HIT posted on Mechanical Turk.}
\end{center}
\end{figure}

Fig.~\ref{hard-sarcasm-examples} provides examples of response posts
that are initially categorized differently by Karger's and by the
threshhold method of \cite{LukinWalker13}.  The {\bf final} columns in
Fig.~\ref{hard-sarcasm-examples} show that different methods often
converge after collecting 25 additional annotations.

\begin{table}[h]
\centering
\begin{tabular}{|c|c|c|c|}
\hline  
 &  \multicolumn{3}{c|}{\bf Karger} \\
{\bf L\&W13} &  \bf AMBIG & \bf NOT-SARC & \bf SARC \\ \hline \hline  
\bf  NOT-SARC  &            47   &            1348   &            8 \\
\bf  SARC       &          109  &             1125  &           172 \\
\hline  
\end{tabular}
\caption{Summary of the relationship between L\&W labels and Karger's labels
on the L\&W13 dataset. \label{karg-steph-crosstab}}
\end{table}

One of our primary motivations is to explore question
{\bf O1}, i.e. what different reliability models indicate
about L\&W's experimental method and results. Table~\ref{karg-steph-crosstab} summarizes the relationship between
L\&W labels and the Karger categorization 
according to the threshhold for Karger's described in
Sec~\ref{method-sec} In L\&W none of the
examples were originally classified as AMBIG. Note the large number of
examples (1125) categorized as SARC by L\&W but categorized as
NOT-SARC by Karger. This is because Karger treats the SARC label and
the NON-SARC label the same, and in many cases the majority of
annotators may have used the NON-SARC label. The assumption behind the
L\&W threshhold (two annotators said it was SARC) is 
that some annotators may be better attuned to 
sarcasm (question {\bf O2}), i.e. that the interpretative
process to recognize sarcasm is similar to 
that for indirect speech acts \cite{Levinson85}.

The Karger AMBIG region contains a large number of posts that L\&W consider
as SARC. 

\begin{figure}[h!]
\begin{center}
\includegraphics[width=3.0in]{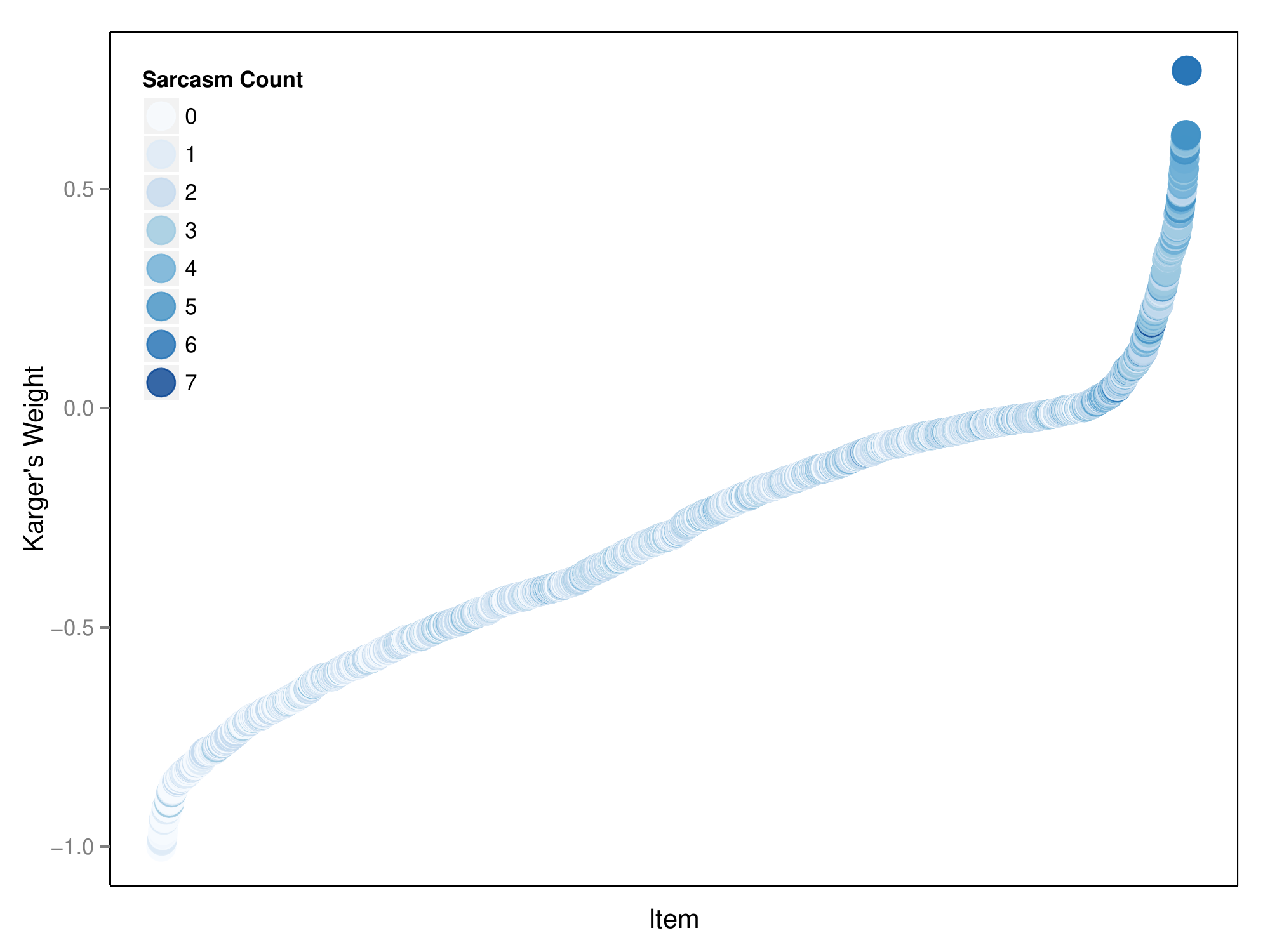}
\vspace{-0.2in}
\caption{\label{sarc-count-karger} Sarcasm annotator counts from
0 to 7 for {\bf L\&W}'s corpus, according to Karger values.}
\end{center}
\end{figure}

Fig.~\ref{sarc-count-karger} shows the relationship between the number
of annotators that marked a post as sarcastic in the original IAC
corpus and Karger scores. Remember that L\&W threshhold was two
annotators. The upper part Fig.~\ref{sarc-count-karger} is deeper blue
as expected, but Karger values below zero were considered NOT-SARC for
the purpose of this experiment, thus Karger values clearly provide a
different threshhold for sarcasm classification, as also illustrated
by Table~\ref{karg-steph-crosstab}.

\begin{figure}[h!]
\begin{center}
\includegraphics[width=3.0in]{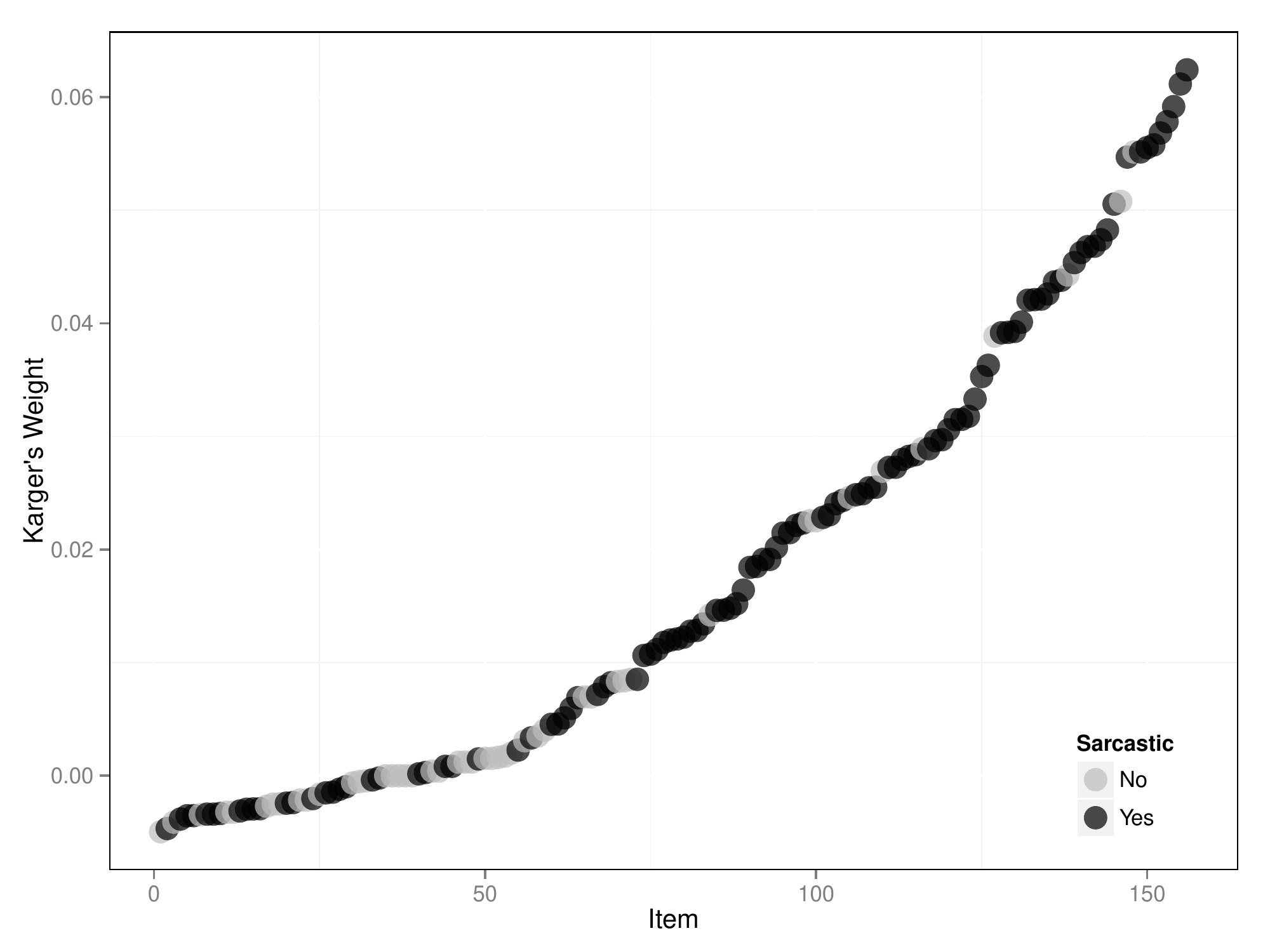}
\vspace{-0.2in}
\caption{\label{karger-original-ambig} Sarcasm and Karger values
for the subset of the {\bf L\&W} corpus that fall within the
area that we consider initially AMBIG using Karger values.}
\end{center}
\end{figure}

Fig.~\ref{karger-original-ambig} explodes the ambiguous portion of
Fig.~\ref{karger-original} to show more clearly how L\&W tokens of
SARC vs. NOT-SARC posts distribute across the ambiguous region
according to their Karger score. While the lower end of the Karger
scale clearly contains more NOT-SARC posts, and the upper end contains
more SARC posts, we observe that the two methods produce very
different values in the AMBIG range. Note that many posts
that Karger's considers ambiguous in 
Fig.~\ref{hard-sarcasm-examples} appear to end up as classified
as SARC.

We then collected annotations for these 600 utterances
from 25 Mechanical Turkers, who were highly
qualified and resident in the U.S. but who are not master Turkers.
Mechanical turkers were shown a series of 10 Q-R pairs, as illustrated
by the example HIT in Fig.~\ref{hit-fig}. To prevent bias, Turkers
were told that some, all, or none of these pairs may contain sarcastic
utterances. However, of the 10 items in each HIT, 2 were selected from
the {\bf 100-SARC} set, 2 were selected from the {\bf 100-NOT-SARC}
set, and the rest were selected from the {\bf 400-AMBIG} set. See
Fig.~\ref{initial-distrib-fig}.  Turkers were asked to indicate on a
binary scale whether they believed that any part of the responses in
these pairs were sarcastic. To ensure that Turkers gave their best
effort to correctly annotate every post, they were not given the
option to annotate posts as ambiguously sarcastic, and were instead
instructed to use their best judgment if they could not tell if a post
utilized sarcasm.

\section{Results}
\label{results-sec}

We use the 25 new annotations to compare the different reliability
measures on our gold standard data in terms of accuracy as a function
of the number of Turker annotations. We also examine the implications
of this study for the L\&W sarcasm data. Finally, we examine the value
of additional annotations and the implications for future similar
studies.

\begin{figure}[h!]
\begin{center}
\includegraphics[width=3.25in]{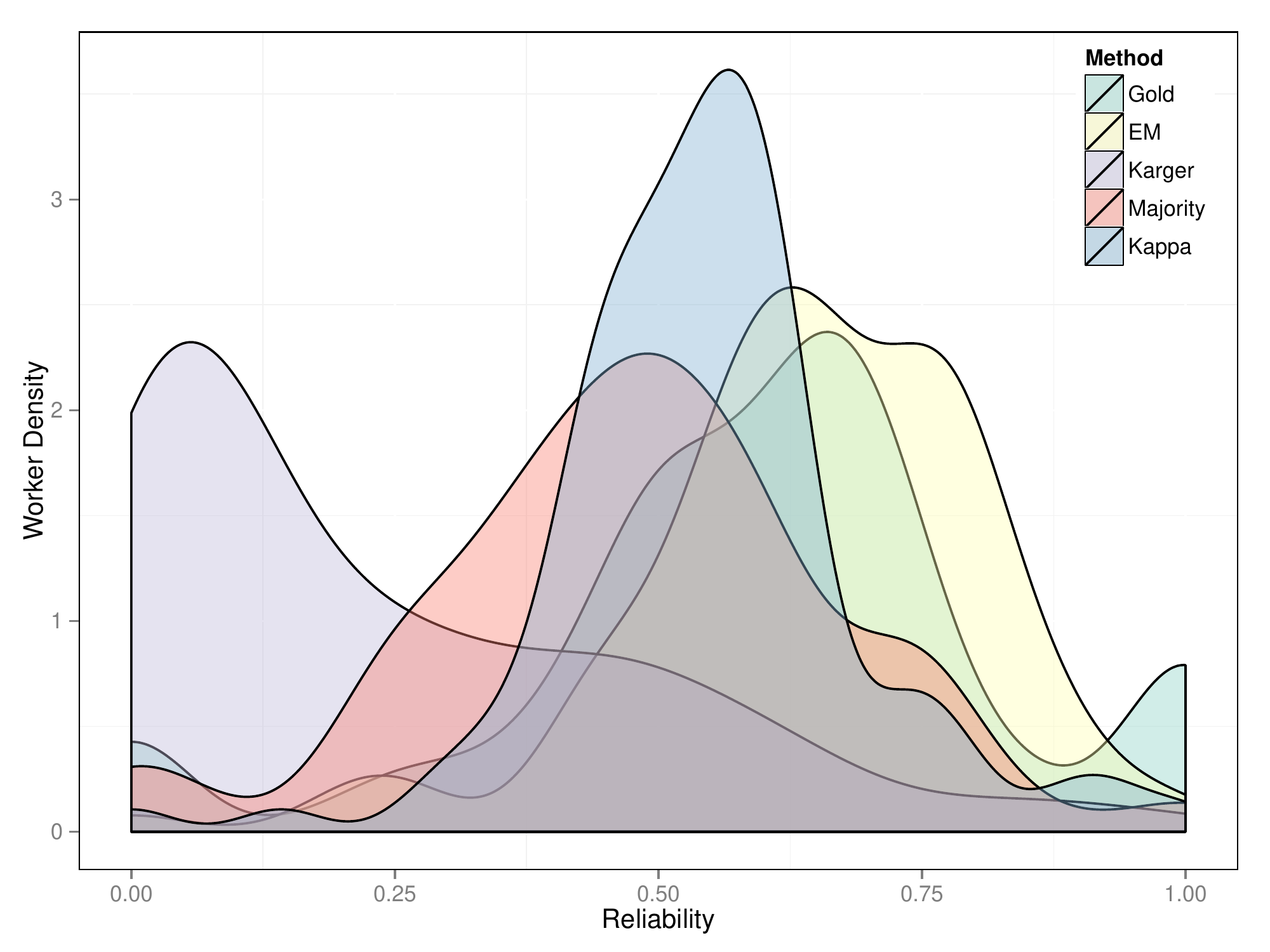}
\vspace{-0.2in}
\caption{\label{reliability-density-fig} The estimated reliability density of workers for each method.}
\end{center}
\end{figure}

\noindent{\bf Reliability Density of Workers.} Fig.~\ref{reliability-density-fig} shows the estimated worker
reliability density for each method normalized between 0 and 1 to ease
the comparison.  The accuracy on the gold standard questions is shown
under the curve in yellow.  Most workers got about $\frac{2}{3}$ of
the gold standard questions correct.  There were slightly more
annotators who fell below the peak, but there were also many who
answered nearly all of them correctly.  This indicates that while
there are many unreliable annotators (i.e., those near 50\%) there are
also many who provide useful answers.

Each of the four reliability measures provides a distinct distribution.
Karger's method tends to think most annotators are unreliable and only assigns a high reliability to the few annotators that performed most of the HITs.
In contrast, the Kappa method places most of the density mass in the middle of the scale with few annotators achieving high or low reliability.
Majority voting is similar to the Kappa method, but has a larger variance.
The EM method tends to over estimate the reliability of the workers, but has the closest distribution to the gold standard.

\begin{figure}[htb!]
\begin{center}
\includegraphics[width=3.0in]{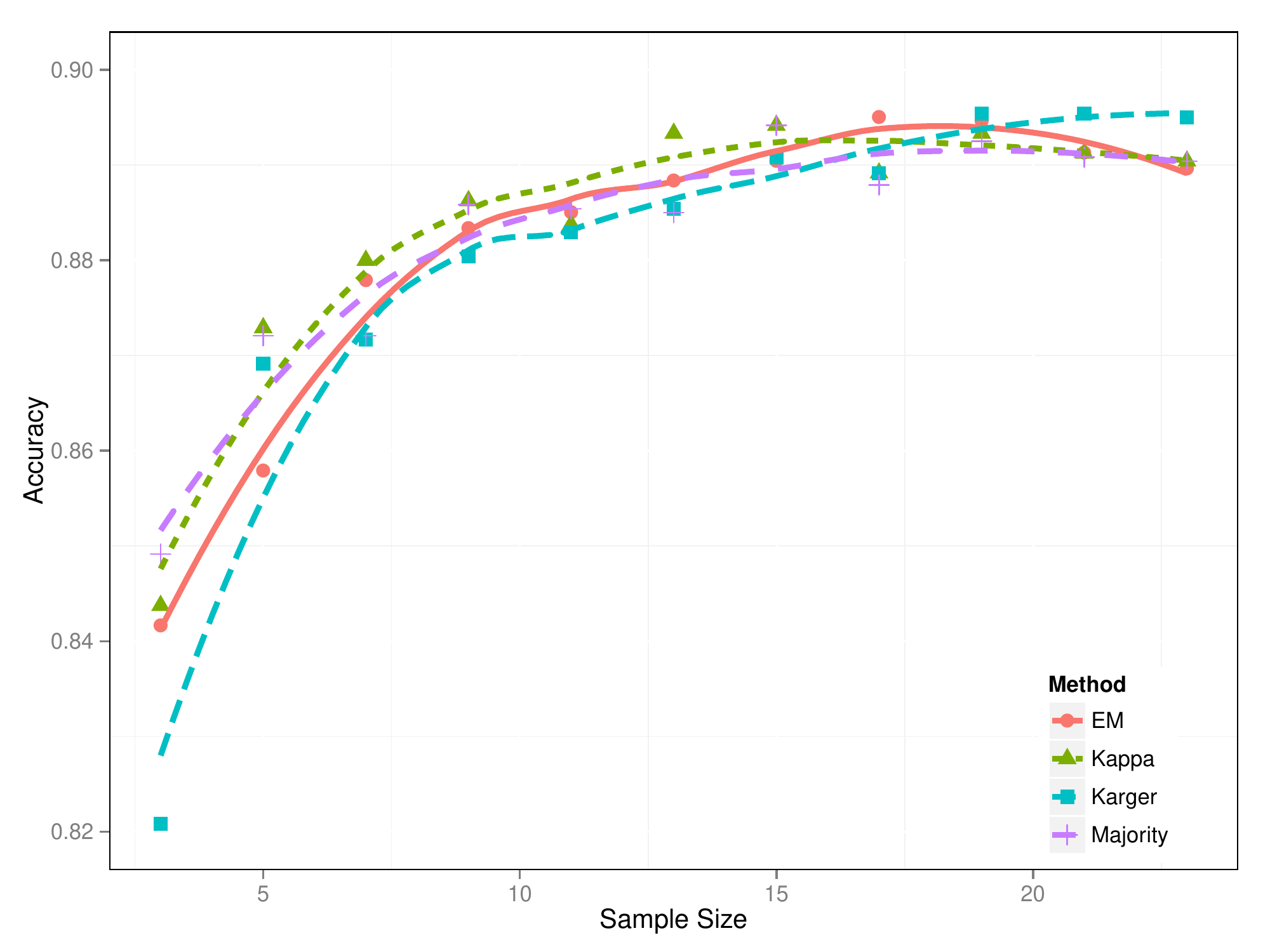}
\vspace{-0.2in}
\caption{\label{reliability-graph} The labeling accuracy of each model compared against the input gold standard (200 of the 600 experimental utterances). For each method
a subset of annotators (from 3-23) was selected to compare the models when fewer annotators are available.
Each subset was run 10 using a random sample from the 25 available annotators for each item. The mean
accuracy is plotted.}
\end{center}
\end{figure}

\begin{figure}[htb!]
\begin{center}
\includegraphics[width=3.0in]{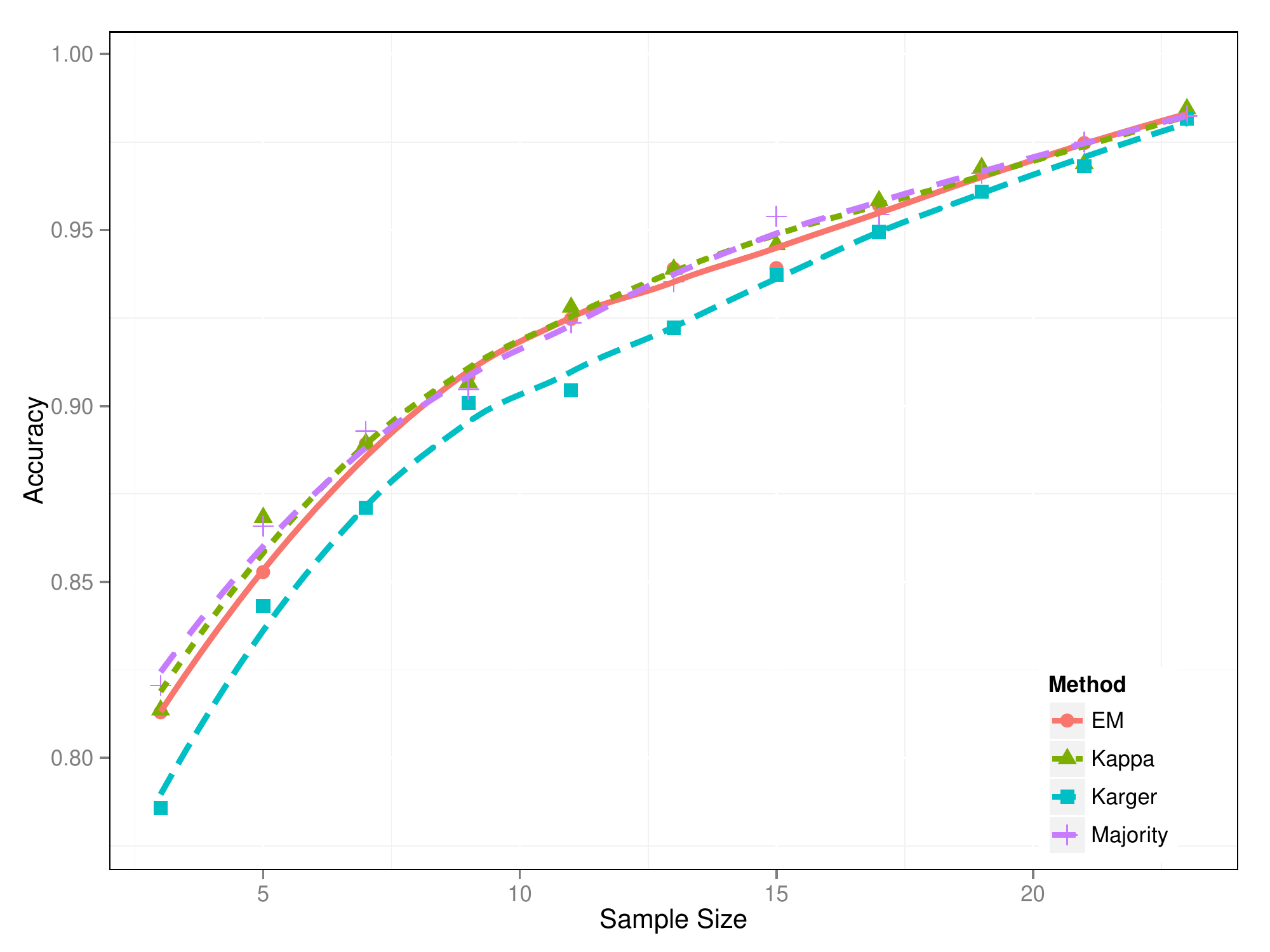}
\vspace{-0.2in}
\caption{\label{reliability-graph-ambig} The labeling accuracy of the 400 
Karger ambiguous cases for each model compared against the final labels assigned by Karger's algorithm with 25 annotators.}
\end{center}
\end{figure}

\begin{figure}[htb]
\begin{center}
\includegraphics[width=3.1in]{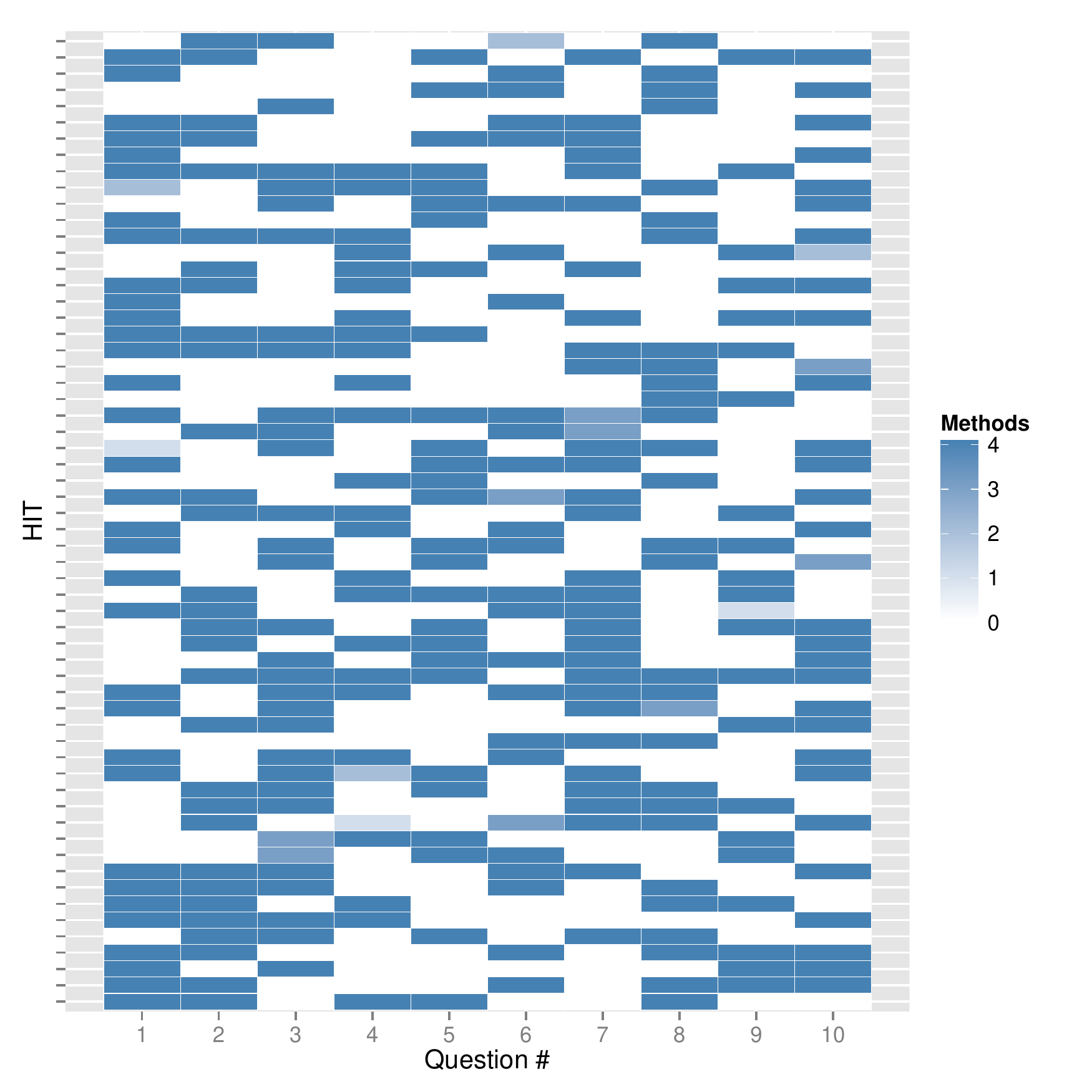}
\vspace{-0.2in}
\caption{\label{final-distrib-fig} 
600 utterances from IAC, shown as ten items per HIT as in the MT study.
This shows the distribution of final labels by multiple voting methods.
White $=$ all methods voted not sarcastic. Deep Blue $=$ all methods
sarcastic. Pale Blue $=$ Still ambiguous, No clear majority
using all 4 methods (10 or fewer of 600 utterances).}
\end{center}
\end{figure}

\begin{figure}[htb]
\begin{center}
\includegraphics[width=3.1in]{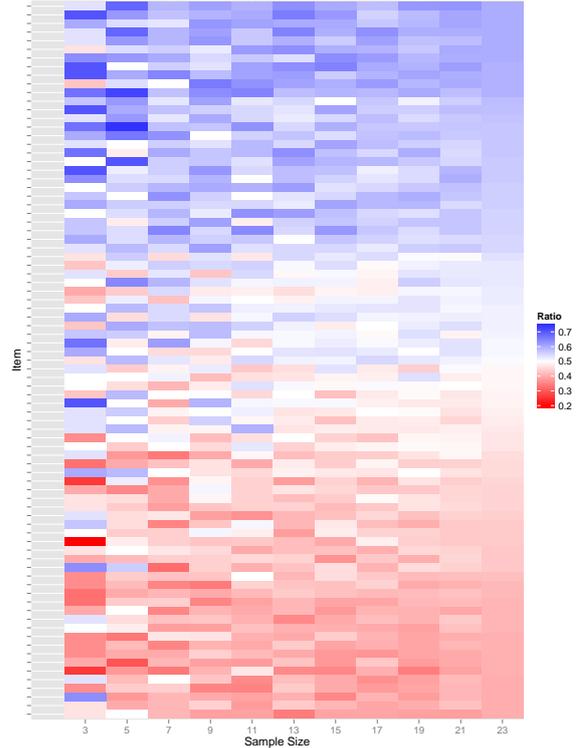}
\vspace{-0.2in}
\caption{\label{unreliable-distrib-fig} 101 items centered around the most ambiguous item (50 above and below) starting with only 3 annotations using Karger's method.}
\end{center}
\end{figure}

\begin{figure}[htb]
\begin{center}
\includegraphics[width=3.1in]{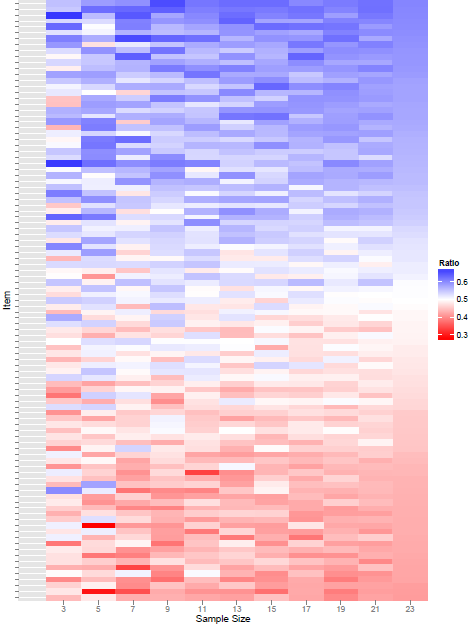}
\vspace{-0.2in}
\caption{\label{unreliable-karger-end} 101 items that were most ambiguous items after 23 annotations using Karger's method.}
\end{center}
\end{figure}

\begin{table}
\centering
\small
\newcolumntype{Z}{>{\raggedleft\arraybackslash}X}
\begin{tabularx}{\columnwidth}{|p{0.3in} r Z Z | r Z Z|}
\toprule
Start & $+ -$ & Mean & Max & $- +$ & Mean & Min \\
\midrule
3     & 18  & 0.53   & 0.60   & 19  & 0.48   & 0.44 \\
5     & 14  & 0.53   & 0.58   & 6   & 0.48   & 0.41 \\
7     & 7   & 0.51   & 0.54   & 10  & 0.47   & 0.45 \\
9     & 10  & 0.52   & 0.56   & 2   & 0.50   & 0.50 \\
11    & 9   & 0.51   & 0.53   & 4   & 0.49   & 0.48 \\
13    & 6   & 0.51   & 0.52   & 5   & 0.48   & 0.47 \\
15    & 7   & 0.51   & 0.54   & 3   & 0.48   & 0.46 \\
\bottomrule
\end{tabularx}
\caption{The number of items that switch labels from sarcastic ($+$) to not sarcastic (-) after 23 annotations given how many annotations you start with. The mean indicates the average ratio of sarcastic to non sarcastic votes based on Karger's method for that label. The max and min values represent largest (or smallest) value in the set (i.e., the most confident in the wrong label).}
\label{tab:switching_items}
\end{table}

\noindent{\bf Labelling Results.} 
Our results are summarized in
Fig.~\ref{reliability-graph}, Fig.~\ref{reliability-graph-ambig} and Fig.~\ref{final-distrib-fig}.
Fig.~\ref{reliability-graph} suggests that {\bf for the gold standard
  data set} that we already have highly reliable data (accuracy is
82.5\%) with only a few annotators.  If we increase the number of annotations to 15 or 20
annotators however we achieve an approximately 5\% increase in
annotation accuracy on the gold standard data. 

However, the story is
different for the 400 utterances that were in the Karger ambiguous category.
Fig.~\ref{reliability-graph-ambig} plots the accuracy of the 400 ambiguous cases when using the labels assigned by Karger's algorithm using all 25 of the annotations.
On this data we also start out with a high level of reliability with only 3 annotators (about 80\%). With ten annotators the accuracy is close to 90\%,
 and all methods are within 2\% of the final accuracy (according
to Karger's) by 23 annotations.

Fig~\ref{final-distrib-fig} shows the complete set of
600 utterances after the full annotation study and counts the methods that labeled that item as sarcastic.
This provides another view on which items are still ambiguous, i.e. the ones our methods can't agree on, after collecting 25 annotations.
Only approximately 9 out of 600 are still ambiguous (pale blue in Fig.~\ref{final-distrib-fig}.  Thus for the subclass
of utterances considered AMBIG initially, we achieve reliable labels
for these with an additional 15 to 20 annotations.  The return for the
15 annotations above 10 is small. Some examples of these types of
utterances are provided in Fig.~\ref{hard-sarcasm-examples}.
Interestingly, there are also utterances ({\bf R4} of
Fig.~\ref{hard-sarcasm-examples}) that remain ambiguous even with 30
annotations. 
Table~\ref{kargfinal-steph-crosstab} further illustrates
that the threshhold we use for Karger's is much more
conservative than L\&W's initial threshhold.

\begin{table}[h]
\centering
\begin{small}
\begin{tabular}{|p{.5in}|p{.5in}|c||p{.5in}|c|}
\hline  
\bf L\&W &  \multicolumn{2}{c||}{\bf INITIAL} &  \multicolumn{2}{c|}{\bf FINAL} \\
\bf SARC COUNT  &  \bf NOT-SARC & \bf SARC  &  \bf NOT-SARC & \bf SARC \\ \hline \hline  
   0 & 22 &  3  &   22 &   3  \\
   1 &  14 &   15 &   23 &   6 \\
   2 &  33 &   45 &   34 &  44 \\
   3 &  31 &   47 &  17 &  61 \\
   4 &  12  &   28 &  10 &  30 \\
   5  &   3  &   7 &  0 &  10 \\
   6  &  0  &   2 &  0  &  2 \\ \hline
TOTAL & 115 & 147 & 106 & 156\\
\hline  
\end{tabular}\caption{The 262 items from  L\&W  with initial sarcastic counts
in IAC, and initial
and final Karger's labels. \label{kargfinal-steph-crosstab}}
\end{small}
\end{table}

To get a better understanding of how many annotations we should
collect in future before reaching diminishing returns, we also
looked at how ambiguous cases changed labels as more annotations were
provided.  Fig.~\ref{unreliable-distrib-fig} shows 101 items that were
the most ambiguous after 3 annotations of Karger's algorithm using the
same sampling method as Fig~\ref{reliability-graph} and then tracks
how the labels change as more annotations are provided.  To enable
comparisons between other methods we defined ambiguous as the items
whose ratio of sarcastic votes to non sarcastic votes was closest to
0.5.  Although many of the items that start out closest to 0.5 change
their initial assignment, most have stabilized by 5-7 annotations and
only a handful continue to oscillate.  Fig~\ref{unreliable-karger-end}
shows a similar plot of the items that were the most ambiguous after
sampling 23 annotations.  This plot also seems to indicate that most
labels have reached their final target after 7 annotations.

Table~\ref{tab:switching_items} provides the exact number of items on
the entire dataset that switched labels after 23 annotations were
sampled based on how many annotations were provided to start with.
The table also shows the mean ratio for the items that switched label
and the maximum (for $+ -$) and minimum (for $- +$) values.  On
average, only the items that start out highly ambiguous with only a
few labels change labels when over 20 annotations are provided.  It
also shows that as the number of starting annotations increases the
mean and max values stabilize around 7 annotations, with a few minor
fluctuations.  Our conclusion from this data is that in most cases we
only need to acquire 3 annotations and can increase the number up to 7
when the initial judgments are highly ambiguous.

\section{Discussion and Future Work}
\label{discuss-sec}

We report the results of a detailed annotation study for sarcasm on
Mechanical Turk using different reliability measures.  Unfortunately
question {\bf O1} still remains open: while only 6 items remain
ambiguous at the end of the study if we apply the voting method, the
overlap of these with the L\&W data is small (262 items). We do not
have the results for final voting on all 600 items in terms of L\&Ws
original threshhold.  We hypothesize that the answer to question {\bf
  O2} is that sarcasm may be more difficult to get reliable
annotations for, and that some utterances may be designed to be
deliberately sarcastic.  However it appears that 7 Turkers are
sufficient to converge on a label category. This means that the
answer to question {\bf O3} is no, because previous work suggested that
for subjective tasks, 7 annotators is enough to achieve reliable
annotations \cite{Snowetal08,callison2009fast}: our results that 
labelling categories for sarcasm converge at around 7 to 10 annotators.

\bibliographystyle{lrec2014}
%\bibliography{../phd,../nl,../all}
%\bibliography{../phd,../nl,../reid}

\end{document}